\documentclass{article}

\usepackage{microtype}
\usepackage{graphicx}
\usepackage{subfigure}
\usepackage{booktabs} 

\usepackage{hyperref}

\usepackage[accepted]{icml2025}

\usepackage{amsmath}
\usepackage{amssymb}
\usepackage{mathtools}
\usepackage{amsthm}

\usepackage[capitalize,noabbrev]{cleveref}

\theoremstyle{plain}

\theoremstyle{definition}

\theoremstyle{remark}

\usepackage[textsize=tiny]{todonotes}

\usepackage{multirow}
\usepackage{xspace}
\newcommand{\ours}{MARASON\xspace}

\icmltitlerunning{Neural Graph Matching Improves Retrieval Augmented Generation in Molecular Machine Learning}

\begin{document}

\twocolumn[
\icmltitle{Neural Graph Matching Improves Retrieval Augmented Generation \\in Molecular Machine Learning}



\icmlsetsymbol{equal}{*}

\begin{icmlauthorlist}
\icmlauthor{Runzhong Wang}{equal,yyy}
\icmlauthor{Rui-Xi Wang}{equal,yyy}
\icmlauthor{Mrunali Manjrekar}{yyy}
\icmlauthor{Connor W. Coley}{yyy}
\end{icmlauthorlist}

\icmlaffiliation{yyy}{Massachusetts Institute of Technology, Cambridge, MA, United States}

\icmlcorrespondingauthor{Connor W. Coley}{ccoley@mit.edu}

\icmlkeywords{Machine Learning, ICML}

\vskip 0.3in
]



\printAffiliationsAndNotice{\icmlEqualContribution} 

\begin{abstract}

Molecular machine learning has gained popularity with the advancements of geometric deep learning. In parallel, retrieval-augmented generation has become a principled approach commonly used with language models. However, the optimal integration of retrieval augmentation into molecular machine learning remains unclear. 
Graph neural networks stand to benefit from clever matching to understand the structural alignment of retrieved molecules to a query molecule.
Neural graph matching offers a compelling solution by explicitly modeling node and edge affinities between two structural graphs while employing a noise-robust, end-to-end neural network to learn affinity metrics. We apply this approach to mass spectrum simulation and introduce \ours, a novel model that incorporates neural graph matching to enhance a fragmentation-based neural network.
Experimental results highlight the effectiveness of our design, with \ours achieving 28\% top-1 accuracy, a substantial improvement over the non-retrieval state-of-the-art accuracy of 19\%. Moreover, \ours outperforms both naive retrieval-augmented generation methods and traditional graph matching approaches. 
Code is publicly available at \url{https://github.com/coleygroup/ms-pred}.

\end{abstract}

\section{Introduction}
Enhancing neural networks with task-relevant factual knowledge has shown great promise in advancing knowledge-intensive applications, a technique widely recognized as retrieval-augmented generation (RAG)~\citep{lewis2020retrieval}. In scientific domains, where the demand for accurate and reliable models is prominent, retrieval-augmented generation has achieved significant success. 
Recent progress in small-molecule research further highlights the potential of retrieval-augmented generation, which is also the main focus of this paper.\footnote{``Small molecules'' are defined in this work by a mass upper limit of 1500 Da, enforced in all experiments.} This approach has been demonstrated to enhance the accuracy and robustness of various molecular machine learning applications, including structure-based drug design~\citep{zhang2024rag2mol,huang2024interaction}, fragment-based drug discovery~\citep{lee2024molecule}, and monomer design for advanced materials~\citep{buehler2024generative}.

We assume a database exists with pairs of molecular structures and their properties of interest. For a new structure of interest, references in the database can be straightforwardly retrieved with cheminformatic tools such as molecular fingerprints~\citep{morgan1965generation} and Tanimoto similarity~\citep{bajusz2015tanimoto}. Domain experts benefit from such features in databases by searching for pairs of structures to correlate differences in structures with differences in properties, even leading to entire subfields like matched molecular pair analysis~\citep{griffen2011matched}. Therefore, we expect that augmenting a structure-property relationship model with structurally similar molecules and their associated properties should improve the accuracy of predictions. 

However, designing an effective neural network architecture for retrieval augmentation presents significant challenges. Simple approaches, such as direct concatenation of reference information to the input of the neural network, often yield minimal or no improvements (see Table~\ref{table:ablation}). We hypothesize that this is due to models' inabilities to adequately address differences between the retrieved references and the target molecule. We propose that employing atom-level or fragment-level matching mechanisms in an end-to-end network would allow for reference information to be used more effectively.


\begin{figure}[tb!]
    \centering
    \includegraphics[width=0.8\linewidth]{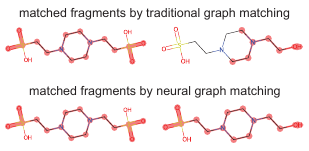}
    \vspace{-10pt}
    \caption{Comparison of graph matching results between a traditional algorithm \citep{ChoECCV10} and the neural graph matching module in this paper. Neural graph matching is more robust especially when the matched structures are not identical. More visualization results can be found in Fig.~\ref{fig:pairs}, \ref{fig:pairs_rrwm} in Appendix.}
    \vspace{-10pt}
    \label{fig:gm-compare}
\end{figure}

Graph matching, which addresses node-level correspondences across multiple graphs, naturally emerges as a suitable approach to this challenge. It explicitly incorporates structural matching by formulating both node-wise and edge-wise graph affinity scores into the quadratic assignment problem~\citep{LawlerMS63}. 
In molecular tasks, graph matching recapitulates what domain experts might do when comparing two molecules. 
In practice, traditional graph matching methods typically define the graph affinity using pre-established metrics, such as Gaussian kernels based on Euclidean distance. Such a predefined affinity metric has a significant limitation: its limited expressivity makes traditional graph matching methods vulnerable to random noise and not robust to possible ambiguities in retrieved structures~\citep{WangPAMI21}.
To address this robustness challenge, a new class of neural graph matching methods has emerged, which learns the affinity metric and the solver module in an end-to-end manner~\citep{ZanfirCVPR18,WangPAMI20,LiICML19}.
As a result, neural graph matching provides a compelling design choice for retrieval augmentation, as shown in Fig.~\ref{fig:gm-compare}. Chemists are able to draw analogies between molecules beyond simple maximum common substructure analysis and consider, for example, the relationship between non-equivalent but functionally similar (i.e., isosteric) functional groups. 
By integrating this chemistry-inspired view of graph matching with the expressivity and adaptability of neural networks, neural graph matching represents an effective framework for aligning target structures with, and thus learning from, their retrieved counterparts.

To this end, we implement and validate our design strategy in a prominent application of molecular machine learning: the neural simulation of spectra from tandem mass spectrometry (MS/MS). MS/MS is an analytical chemistry technique that generates profiles from unknown molecules which are used as diagnostic signatures in structure elucidation workflows. This approach has wide-ranging applications in chemistry and biology, including biomarker discovery~\citep{dang2009cancer}, metabolomics~\citep{quinn2020global}, and environmental science~\citep{tian2021ubiquitous}, among others. An MS/MS simulator is designed to take molecular structures as input and predict their mass spectra, i.e., a set of mass-over-charge ($m/z$) values and peak intensities. Simulating these spectra can accelerate the structural elucidation pipeline, illuminating the spaces of so-called ``metabolite dark matter''~\citep{bittremieux2022critical}. 

\begin{figure*}
    \centering
    \includegraphics[width=\linewidth]{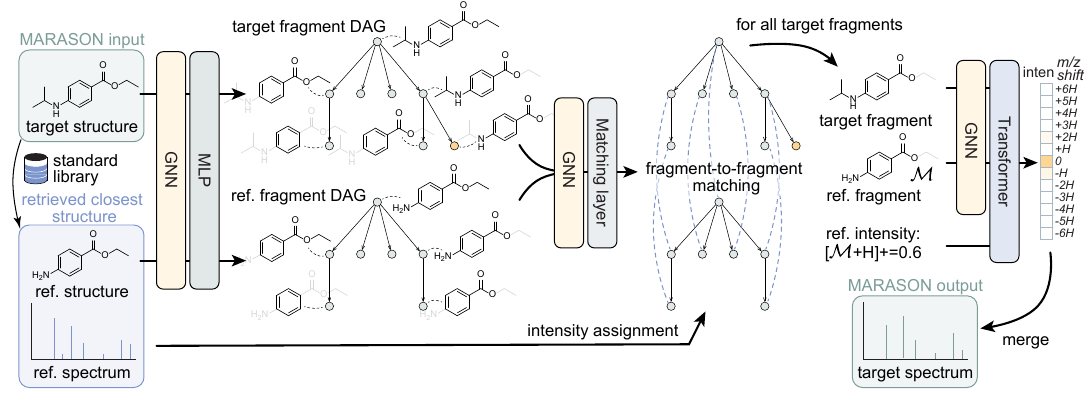}
    \vspace{-20pt}
    \caption{Overview of \ours: a retrieval-augmented mass spectrum simulator with neural graph matching. We retrieve reference structures and spectra from a reference library (more specifically, the training dataset) based on Tanimoto similarity~\citep{bajusz2015tanimoto} to the target structure. Both target and reference structures are fragmented, resulting in two similar fragmentation DAGs, 
    as similar structures are expected to have similar fragmentation patterns in chemistry~\citep{shahneh2024modifinder}. The neural graph matching module further finds the node-level matching between fragmentation DAGs, whereby each node represents a fragment. The aligned target and reference fragment, together with reference intensity, are further concatenated to predict the final spectrum.We use three identical GNN modules with separate weights to learn embeddings: one for the target fragments, one for the reference fragments, and one shared by both target and reference fragments to capture matching information. After computing the embeddings with these GNN modules, we use the shared GNN’s outputs as input to a matching module, which produces a matching matrix. This matrix is then used to align reference fragments and spectral peaks with the target fragments. The resulting matched fragment pairs, along with their similarity scores, are fed into a transformer module to predict the final relative intensities of the spectral peaks corresponding to each target fragment.}
    \label{fig:overview}
\end{figure*}

\textbf{Our contributions are:}

\textbf{1)} We present an effective design strategy for retrieval-augmented generation in molecular machine learning with neural graph matching. Graph matching techniques incorporate both node-wise and edge-wise affinities explicitly, providing a principled way of assigning node-to-node or fragment-to-fragment alignment between the reference structure and the target structure\footnote{In this paper, we refer to all structures and spectra retrieved from the database as ``reference'', and the structures and spectra of interest as ``target''.} that emulates how domain experts leverage retrieved references. By further introducing learning-based affinity metrics and differentiable matching layers, we improve the overall efficacy of matching and the resulting accuracy in applications.

\textbf{2)} We present \underline{m}atching-\underline{a}ware \underline{r}etrieval \underline{a}ugmented \underline{s}pectrometry \underline{o}racle with \underline{n}eural networks (\ours), an implementation of neural graph matching-based retrieval-augmented generator for mass spectra, as shown in Fig.~\ref{fig:overview}.
\ours is built upon the 
ICEBERG~\citep{goldman2024iceberg} model and incorporates design principles of neural graph matching~\citep{WangPAMI21,YuICLR20}. 
In \ours, we retrieve structures with known reference spectra from the training dataset~\citep{nist_database}. We exploit ICEBERG's fragmentation model to annotate reference spectrum peaks with fragments and form a fragmentation directed acyclic graph (DAG) with peak annotations for both the target and reference structures. We then match reference fragments to target fragments to inform the second-stage model's intensity prediction. We incorporate nested graph neural networks (GNNs)~\citep{zhang2021nested,scarselli2008graph} to encode (a)~structural information of each fragment and (b)~hierarchical information of the fragmentation DAG into node embeddings, building a neural graph matching network for MS/MS simulation.

\textbf{3)} Our experimental evaluation on standard benchmarks demonstrates state-of-the-art accuracy on the mass spectrum simulation task, outperforming both RAG and non-RAG baselines and thus validating the effectiveness of our design strategy. Specifically, on the \citet{nist_database} dataset, we improve the top-1 retrieval accuracy from 19\% to 28\%. Ablations demonstrate that \ours outperforms a naive RAG baseline (by concatenating the retrieved spectrum to the model input) as well as other matching methods using traditional graph matching solver~\citep{ChoECCV10} or a simple linear matching solver~\citep{Kuhn1955Hungarian}, emphasizing the value of its differentiable, end-to-end neural graph matching module.
Given the broad applicability of mass spectrometry in biological and chemical campaigns, the improved accuracy of our model lends great potential for accelerating molecular discovery by expanding standard mass spectrum libraries with simulated spectra for novel annotation, from a current size of $\sim$27K unique compounds in \citet{nist_database} to all 111M compounds in PubChem and beyond.

\section{Related Work}
\textbf{Retrieval Augmented Generation in AI for Science}.
Retrieval augmented generation~(RAG) is a technique that integrates relevant information retrieved from external databases into a model's training and inference processes. While originally developed to aid Large Language Model (LLM)-based AI agents, their applicability has readily extended to scientific tasks, spanning LLMs for material discovery~\citep{buehler2024generative}, drug design~\citep{LLMDrug}, and organic synthesis~\citep{ChemLLM}. Utilizing RAG in various forms  has also shown ground-breaking advances in scientific discovery beyond LLM agents; one of the most prominent examples is AlphaFold model~\citep{jumper2021alphafold}, whose sequence alignment module is effectively a RAG module. 
RAG is also found helpful in molecular machine learning tasks such as molecular generation~\citep{lee2024molecule}, drug design~\citep{zhang2024rag2mol}, and protein function prediction~\citep{shaw2024protex}. Despite success in certain molecular learning tasks, there remains the opportunity for a principled RAG design strategy tailored to small molecules 
that can perform robust and informative structural matching.

\textbf{Neural Graph Matching}.
Graph matching is a combinatorial optimization problem that matches the nodes of multiple graphs by maximizing the edge-wise and node-wise affinities. Neural graph matching was developed to tackle the computational challenges of the NP-hard quadratic assignment problem~\citep{LawlerMS63} and the vulnerability of predefined affinity metrics~\citep{RolinekECCV20,NowakDSW18,guo2023deep}.
Among all design choices, a family of linear-matching  methodologies~\citep{WangPAMI20,YuICLR20,sarlin2020superglue} incorporates graph neural networks~\citep{scarselli2008graph} to embed edge affinities into node embeddings that are then used in a differentiable node-matching layer using, e.g., \citet{SinkhornAMS64} or a simpler Softmax for assignment. 
This paper explores and validates the applicability of neural graph matching in the context of retrieval-augmented mass spectra generation in molecular machine learning. 

\section{Methods}
\ours is a mass spectrum simulator built on the ICEBERG model~\citep{goldman2024iceberg} with reference retrieval augmentation and neural graph matching.
In this section, unless otherwise specified, we use capitalized bold letters for matrices, lowercase bold letters for vectors, and non-bold letters for scalars. $\mathcal{M}, \mathcal{F}, \mathcal{G}$ indicate a molecular graph, fragment graph, and fragmentation DAG (from ICEBERG-Generate), respectively. All reference-related variables are annotated with a superscript ``$r$''.

\subsection{Preliminary: ICEBERG MS/MS Simulator}
\label{sec:preliminary}
Mass spectrometry (MS) is a powerful analytical chemistry method for the discovery of unknown compounds and natural products~\citep{wang2016gnps}. Compounds analyzed with tandem MS undergo an ionization and fragmentation process, after which charged fragments are detected at their $m/z$ values with an intensity proportional to their relative abundance and cross section. 
\ours is built on the ICEBERG model which reported state-of-the-art accuracy. As described by \citet{goldman2024iceberg}, ICEBERG is a two-stage model:

\textbf{ICEBERG-Generate} learns to mimic collision-induced dissociation by predicting the most likely fragments through recursive removal of atoms and bonds from the molecular graph. It results in an autoregressive model to handle multiple breakages, whose output is a directed acyclic graph (DAG) where the root node is the input molecule and fragments are the children. The hierarchical information of fragments from ICEBERG-Generate is used in \ours for graph matching.

\textbf{ICEBERG-Score} takes the fragmentation DAG and outputs the intensities of peaks corresponding to each fragment. As shown in Fig.~\ref{fig:overview}, the output layer also accounts for up to $\pm$6 hydrogen shifts in mass to support chemical rearrangements. ICEBERG-Score is composed of 1) a GNN module that takes in the molecular graphs and context parameters and outputs the corresponding hidden representation of fragments and root molecules; and 2) a set transformer~\citep{lee2019set} that predicts intensity values for each fragment. Predicted intensity values are merged into a mass spectrum, where the $m/z$ values are straightforwardly calculated from the structures of fragments.

\ours aims to improve ICEBERG by integrating retrieval augmentation with neural graph matching into ICEBERG-Score. 

\subsection{Retrieval Augmentation Processing}
\label{sec:rag}
With the chemical intuition that structurally similar molecules are expected to have similar spectra~\citep{shahneh2024modifinder}, we retrieve molecules with the highest structural similarity to our target molecule. This subsection presents our implementation of MS/MS retrieval and processing of retrieved data for best deep-learning efficacy.
\subsubsection{Retrieval of Structures and Spectra}
\label{sec:retrieval}
Let our retrieval database be any database with annotated structure-spectrum pairs; in our experiments, we use the training dataset to mitigate concerns about data leakage. We retrieve the most similar reference molecule as quantified by Tanimoto similarity~\citep{bajusz2015tanimoto} operating on Morgan fingerprints~\citep{morgan1965generation}. 
We exclude the entries that have nonmatching adduct types or instrument types recorded to mitigate the confounding effects of different experimental conditions. 

We then identify up to three reference spectra with the most similar collision energy to the target (query) structure $\mathcal{M}$. 
Collision energies influence the extent of fragmentation and shift the spectrum patterns towards lower m/z ranges with higher energies; there are often multiple spectra with different collision energies for a unique molecule in \citet{nist_database}. 
The result of retrieval is therefore a single reference structure $\mathcal{M}^r$, $\mathrm{Tanimoto}(\mathcal{M}, \mathcal{M}^r)$, and up to three reference spectra and their collision energy values, which are fed into \ours.
As the collision energy value is continuous, our aim of including three reference energies is to learn an appropriate spectrum embedding for the target collision energy value by interpolating from three closest energies.

\subsubsection{Peak-Fragment Assignment and Learning}
\label{sec:peak-frag-align}
Reference spectra only provide $m/z$ values and peak intensities. This information is enriched by annotating peaks with ICEBERG-predicted fragments of the reference structure. 
For each fragment $\mathcal{F}_j^r$, we assign all peaks that fall within the 20 ppm (parts-per-million) range of at most 13 mass values: $\{\mathcal{F}^r_j-6\delta, \mathcal{F}^r_j-5\delta, ..., \mathcal{F}^r_j, ..., \mathcal{F}^r_j+6\delta\}$, where $\delta$ is the mass of a hydrogen atom. Such a process leads to a 13-dimensional vector for each  $\mathcal{F}_j^r$ indicating its intensities in the reference spectrum. 
The assigned 13-dimension intensity is concatenated with values of the reference and target collision energies. 
All reference intensities at the same collision energy are processed by a set transformer, followed by an average pooling layer that merges intensity embeddings per fragment from three collision energies. The obtained reference intensity embeddings, denoted as $\mathbf{T}^r$, are defined such that the pooled embedding for fragment $j$ is represented as $\mathbf{t}_j^r$.

\subsection{Fragmentation DAG Graph Matching}
\label{sec:method-gm}
We define a fragmentation DAG graph matching problem to recapitulate domain experts' practice of comparing similar fragments from the reference and the target, alongside their hierarchy with respect to the original structures. 
In fragmentation DAGs from the ICEBERG-Generate model, fragments are viewed as nodes, and fragmentation paths are viewed as edges. Since each fragment is a molecular graph, a fragmentation DAG is a graph of graphs, i.e., a meta-graph. Following training steps in \citet{goldman2024iceberg}, a pretrained ICEBERG-Generate model predicts two fragmentation DAGs: one for the reference molecular graph and one for the target.  We then approach the graph matching task using fixed affinity metrics with traditional graph matching solvers as well as learnable affinity metrics with neural graph matching. 

\subsubsection{Traditional Solvers with Fixed Metrics}
\label{sec:method-traditional-gm}
Defining the affinity metric is the first step of matching fragments between two fragmentation DAGs. Since each fragment is a molecular graph, we resort to Tanimoto similarity~\citep{bajusz2015tanimoto} and construct the following pairwise affinity matrix $\mathbf{M}$:
\begin{equation}
    m_{i,j} \gets \mathrm{Tanimoto}(\mathcal{F}_i, \mathcal{F}_j^r),
\end{equation}
where $\mathcal{F}_i$ is fragment $i$ of the target structure and $\mathcal{F}_j^r$ is fragment $j$ of the reference structure. With the Tanimoto-based fragment-level affinity scores, we can formulate a linear assignment problem:
\begin{equation}
\begin{split}
    &\max_\mathbf{X} \ \mathrm{tr}(\mathbf{M}^\top\mathbf{X}), \\
    s.t. \quad &\mathbf{X}\in\{0,1\}^{n\times n^r}, \mathbf{X}\mathbf{1}_{n^r} \leq \mathbf{1}_{n}, \mathbf{X}^\top\mathbf{1}_{n} \leq \mathbf{1}_{n^r},
    \label{eq:lap}
\end{split}    
\end{equation}
where $\mathbf{X}$ is the matching matrix, $n$ and $n^r$ are the number of fragments from the target DAG and the reference DAG, respectively, $\mathbf{1}_n$ means a column vector of $n$ 1s. $x_{i,j}=1$ means $\mathcal{F}_i$ is matched to $\mathcal{F}_j^r$ and $x_{i,j}=0$ otherwise. Eq.~(\ref{eq:lap}) can be solved by the Hungarian algorithm~\cite{Kuhn1955Hungarian}, whereby it incorporates fragment-level graph affinities but ignores the hierarchical information in DAGs. 

Graph matching further extends Eq.~(\ref{eq:lap}), by explicitly modeling edge affinities in the DAG, resulting in a quadratic assignment problem~\citep{LawlerMS63}:
\begin{equation}
    \max_\mathbf{X} \ \text{vec}(\mathbf{X})^\top\  \mathbf{K} \ \text{vec}(\mathbf{X}), \\
    \label{eq:qap}
\end{equation}
where the constraints are the same as Eq.~(\ref{eq:lap}), $\mathbf{K}\in \mathbb{R}^{nn^r\times nn^r}$ is the affinity matrix, and $\text{vec}(\cdot)$ means column-wise vectorization. The diagonal elements of $\mathbf{K}$ are taken to be $\mathbf{M}$ as the fragment-level affinities, while the off-diagonal elements are constructed by the inner product of edges of DAGs. Solving Eq.~(\ref{eq:qap}) is NP-hard, but there are approximate solvers available~\citep{ChoECCV10,LeordeanuNIPS09} through a Python interface~\citep{wang2024pygmtools}. 

After solving Eq.~(\ref{eq:lap}) or Eq.~(\ref{eq:qap}), we establish the matching between fragments and retrieved intensities using the matching result $\mathbf{X}$. As shown in Fig.~\ref{fig:gm-compare} and later ablation studies (Table~\ref{table:ablation}), these methodologies yield inferior results to neural graph matching, indicating the importance of flexible, learnable affinity metrics.

\subsubsection{Neural Matching with Learned Metrics}
\label{sec:method-ngm}
The biggest drawback of traditional graph matching methods is that their fixed affinity metrics do not incorporate the flexibility required to handle noisy real-world data~\citep{WangPAMI20,ChoICCV13}.
To overcome this challenge in the context of RAG for MS/MS simulation, we propose a neural graph matching method that leverages message-passing networks on the fragmentation DAG. Specifically, we develop a nested GNN~\citep{zhang2021nested} that learns fragment-level embedding and DAG hierarchical embedding by two GNNs, followed by a differentiable matching layer.

\textbf{Fragment-level embedding learning}. 
For each fragment $\mathcal{F}_i$ from the target molecule $\mathcal{M}$, since both $\mathcal{F}_i$ and $\mathcal{M}$ are molecular graphs, we learn graph-level embeddings 
with a shared $\mathrm{GNN}_{\text{frag}}$. 
We build embeddings at the fragment level for $\mathcal{F}_i$ using an MLP to project the concatenation of $\mathrm{GNN}_{\text{frag}}(\mathcal{M})$, $\mathrm{GNN}_{\text{frag}}(\mathcal{F}_i)$, and $\left(\mathrm{GNN}_{\text{frag}}(\mathcal{M})-\mathrm{GNN}_{\text{frag}}(\mathcal{F}_i)\right)$, together with an encoded number of broken bonds (from ICEBERG-Generate), encoded chemical formula of $\mathcal{F}_i$ and the chemical formula difference from $\mathcal{M}$ to $\mathcal{F}_i$. These differences in formula and $\mathrm{GNN}_\text{frag}$ embeddings represent the chemical concept of ``neutral losses''. The same neural networks are applied to the target fragments and the reference fragments, yielding $\mathbf{H}$ and $\mathbf{H}^r$ respectively as the fragment-level embeddings learned from GNNs.

\textbf{DAG hierarchical embedding learning}. For each fragmentation DAG, we construct two graphs $\mathcal{G}$ and $\mathcal{G}^{-1}$, where $\mathcal{G}$ is the top-down fragmentation DAG in which $e_{ij} \in \mathcal{G}$ if and only if fragment $i$ is a parent of fragment $j$. $\mathcal{G}^{-1}$ is the bottom-up DAG where all edges are reversed. With $\mathbf{H}$ (or $\mathbf{H}^r$) extracted by aforementioned fragment-level GNNs, it is updated as 
\begin{equation}
    \bar{\mathbf{H}} \gets \mathbf{H} + \mathrm{GNN}_\text{fwd}(\mathbf{H}, \mathcal{G}) + \mathrm{GNN}_\text{rev}(\mathbf{H}, \mathcal{G}^{-1}),
\end{equation}
so that the DAG structures are embedded into fragment-level embeddings. We apply the same GNNs for both target DAG and reference DAG. 

\textbf{Similarity and differentiable matching}. We then calculate the similarity matrix $\bar{\mathbf{M}}$ of a given reference target pair as 
\begin{equation}
    \bar{m}_{i,j} \gets \mathrm{cosine}(\bar{\mathbf{h}}_i, \bar{\mathbf{h}}_j^r),
\end{equation}
where $\bar{\mathbf{h}}_i$ is the embedding of fragment $i$ of the target, and $\bar{\mathbf{h}}_j^r$ is the embedding of fragment $j$ of the reference. We can then calculate the matching matrix $\bar{\mathbf{X}}$ as follows
\begin{equation}
    \bar{\mathbf{X}} \gets \mathrm{matching}(\bar{\mathbf{M}}).
\end{equation}
In neural graph matching research, $\mathrm{matching}(\cdot)$ could either be the Sinkhorn~\cite{SinkhornAMS64} or Softmax algorithms; Sinkhorn is preferred in smaller-sized graphs to enforce stronger matching constraints~\citep{WangPAMI20}, whereas Softmax is used in larger-sized graphs for its efficiency~\citep{sarlin2020superglue}. We implement Softmax-based matching in this paper after a careful ablation study, as empirical experiments with $\sim$100 fragments generated by ICEBERG-Generate suggest this number does not pose a serious concern of constraint violation. It is worth noting that $\bar{\mathbf{X}}$ is a continuous matrix so that the neural graph matching pipeline is differentiable, and that $\bar{\mathbf{X}}$ is further integrated into the intensity prediction module so that it will receive gradients during training and learn the affinity metric end-to-end.

\subsection{Intensity Prediction}
\label{sec:inten-pred}
Given the target fragment embeddings $\mathbf{H}$ and the reference fragment embeddings $\mathbf{H}^r$, together with peak intensities $\mathbf{T}^r$ assigned to reference fragments from Sec.~\ref{sec:peak-frag-align}, as well as the matching matrix $\bar{\mathbf{X}}$ bridging target and reference fragments from Sec.~\ref{sec:method-ngm}, we can then create triplets of (target fragments, reference fragments, reference intensities) to serve as input to the intensity prediction layers. To reiterate our motivation for this retrieval-augmented generation framework: similar structures tend to have similar fragmentation patterns in chemistry, and similar fragments tend to have similar response factors that relate their abundance to the observed intensity~\citep{shahneh2024modifinder}. Therefore, the intensity of reference fragments should offer hints for the intensity of the target fragments.

When predicting the intensities, we also consider the matching score of target fragment $i$ as $s_i$:
$
    s_i \gets \sum_{j=1}^{n^r}\bar{x}_{i,j} \bar{m}_{i,j},
$
which is further aggregated with $\mathbf{H}$, $\mathbf{H}^r$, $\mathbf{T}^r$, and the Tanimoto similarity of the original molecules for intensity prediction. The input is the concatenation of
\begin{equation}
    \left[\mathbf{H}, \bar{\mathbf{X}}\mathbf{H}^r, \bar{\mathbf{X}}\mathbf{T}^r, \mathbf{s}, \mathrm{Tanimoto}(\mathcal{M}, \mathcal{M}^r)\right],
    \label{eq:inten-set-trans-input}
\end{equation}
which is then processed by a set transformer~\citep{lee2019set} where embeddings associated with each fragment are treated as one element in the set. Finally, the intensities are computed through an attention layer and an MLP.

\section{Experiments}
We implement \ours with PyTorch~\citep{Paszke2017AutomaticPytorch}, based on the official implementation of ICEBERG~\citep{goldman2024iceberg} and the graph matching toolkit Pygmtools~\citep{wang2024pygmtools}. All experiments are conducted on a workstation with AMD 3995WX CPU, $4\times$NVIDIA A5000 GPU, and 512GB RAM. 
Our experiments are designed to answer the following questions: 
{(1)}~Does \ours, a representative implementation of neural graph matching-based RAG in molecular machine learning, improve the performance of mass spectrum prediction compared to retrieval-free ICEBERG and other state-of-the-art models?
{(2)}~How accurate is \ours on an authentic assessment of identifying unknown compounds to support its application to real-world chemical and biological campaigns?
{(3)}~Does \ours generate a reasonable matching pattern that aligns with the fragmentation process?
We conduct experiments and answer these research questions quantitatively and qualitatively.

\begin{figure}[tb!]
    \centering
    \includegraphics[width=\linewidth]{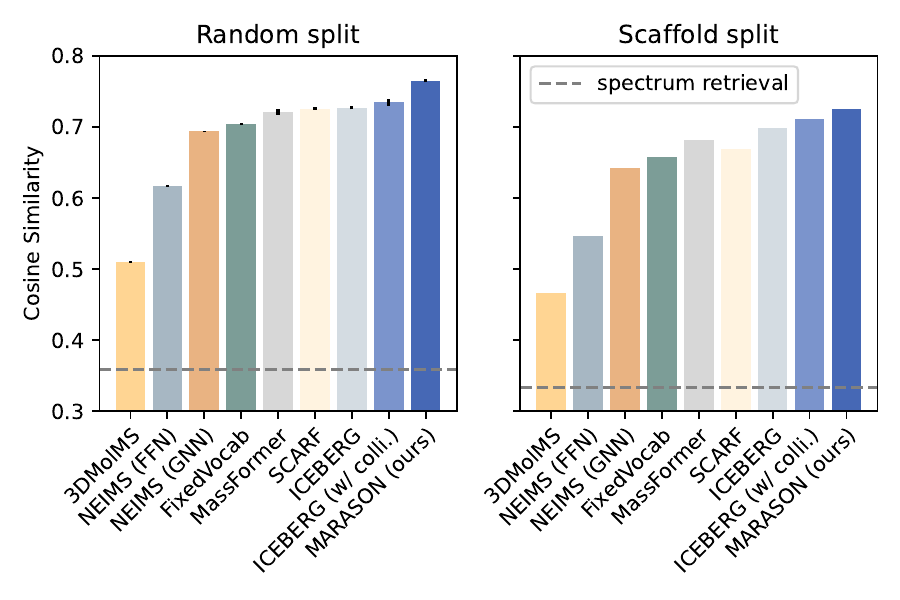}
    \vspace{-25pt}
    \caption{Spectral similarity between experimental spectra and predictions from various MS/MS simulators on \citet{nist_database} dataset. Results of all baselines are from the benchmark developed by \citet{goldman2024iceberg}. Incorporating collision energies with ICEBERG slightly improves the accuracy, while \ours further improves the accuracy by a substantial margin. Error bars are reported for the random split with 3 random seeds. The scaffold split is more challenging and results in lower cosine similarity for all methods. It is worth noting, from the dashed lines in the plots, that the retrieved spectra are informative, but the cosine similarities of the retrieved spectra are lower than the predicted spectra produced by all baseline models, demonstrating that RAG on MS/MS is a non-trivial task.}
    \vspace{-10pt}
    \label{fig:cosine}
\end{figure}

\begin{table*}[tb!]
    \centering
    \caption{Retrieval accuracy (mean $\pm$ 95\% confidence interval on 3 random seeds) on \citet{nist_database} dataset with random (InChI key) split on positive adduct types. \ours surpasses all baselines in terms of retrieval accuracy, which is an authentic assessment of MS/MS simulators' performance in real-world applications where they are used to distinguish the true structure from a list of candidates. Results of all baselines are from the benchmark developed by \citet{goldman2024iceberg}. We also outperform the recently developed method FraGNNet~\citep{young2024fragnnet}, which is reported separately in Table~\ref{table:retrieval_H} in the Appendix as it only allows prediction for [M+H]+ adducts.}
    \resizebox{\linewidth}{!}{
    \begin{tabular}{l|ccccccc}
    \toprule
        Accuracy @ Top-$k$ & 1 & 2 & 3  & 4 & 5 & 8 & 10 \\ \midrule
        3DMolMS~\citep{hong20233dmolms} & 0.055$\pm$0.003 & 0.105$\pm$0.000 & 0.146$\pm$0.005 & 0.185$\pm$0.007 & 0.225$\pm$0.009 & 0.332$\pm$0.005 & 0.394$\pm$0.008 \\ 
        FixedVocab~\citep{murphy2023efficiently} & 0.172$\pm$0.004 & 0.304$\pm$0.004 & 0.399$\pm$0.002 & 0.466$\pm$0.007 & 0.522$\pm$0.012 & 0.638$\pm$0.009 & 0.688$\pm$0.006 \\ 
        NEIMS (FFN)~\citep{wei2019neims} & 0.105$\pm$0.003 & 0.243$\pm$0.012 & 0.324$\pm$0.013 & 0.387$\pm$0.011 & 0.440$\pm$0.014 & 0.549$\pm$0.010 & 0.607$\pm$0.005 \\ 
        NEIMS (GNN)~\citep{zhu2020using} & 0.175$\pm$0.005 & 0.305$\pm$0.003 & 0.398$\pm$0.002 & 0.462$\pm$0.004 & 0.515$\pm$0.005 & 0.632$\pm$0.007 & 0.687$\pm$0.005 \\ 
        MassFormer~\citep{young2024massformer} & 0.191$\pm$0.008 & 0.328$\pm$0.006 & 0.422$\pm$0.004 & 0.491$\pm$0.002 & 0.550$\pm$0.005 & 0.662$\pm$0.005 & 0.716$\pm$0.003 \\ 
        SCARF~\citep{goldman2023scarf} & 0.187$\pm$0.008 & 0.321$\pm$0.006 & 0.417$\pm$0.007 & 0.486$\pm$0.008 & 0.541$\pm$0.009 & 0.652$\pm$0.008 & 0.708$\pm$0.009 \\ 
        \midrule
        ICEBERG~\citep{goldman2024iceberg} & 0.189$\pm$0.012 & 0.375$\pm$0.005 & 0.489$\pm$0.007 & 0.567$\pm$0.005 & 0.623$\pm$0.004 & 0.725$\pm$0.003 & 0.770$\pm$0.002 \\
        ICEBERG (w/ collion energy) & 0.202$\pm$0.009 & 0.399$\pm$0.008 & 0.513$\pm$0.008 & 0.585$\pm$0.008 & 0.639$\pm$0.010 & 0.749$\pm$0.006 & 0.793$\pm$0.007 \\
        \ours (ours) & \textbf{0.278$\pm$0.002} & \textbf{0.455$\pm$0.004} & \textbf{0.562$\pm$0.009} & \textbf{0.636$\pm$0.006} & \textbf{0.685$\pm$0.004} & \textbf{0.784$\pm$0.002} & \textbf{0.827$\pm$0.004} \\
        \bottomrule
    \end{tabular}
    }
    \vspace{-5pt}
    \label{table:retrieval}
\end{table*}

\subsection{Experimental Setup}
\subsubsection{Dataset and Baselines}
We trained our models on the \citet{nist_database} dataset with 530,640 high-energy collision-induced dissociation (HCD) spectra and 25,541 unique molecular structures. The number of spectra is larger than unique structures because spectra are collected at different collision energies with different adducts. The dataset is split into structurally disjoint 80\%-10\%-10\% train-validate-test subsets. 
Following \citet{goldman2024iceberg}, we evaluate on two different splits: (1)~a random split that splits different InChI keys and (2)~a Murcko scaffold split that clusters different molecular scaffolds that require more generalization to out-of-distribution structures.
It is worth noting our dataset is preprocessed differently from \citet{goldman2024iceberg} as we include collision energies as input that are crucial to retrieving the most informative spectra, whereas the original dataset pre-average spectra when multiple collision energies are used. 
We also include negative adduct types in training to exploit the full \citet{nist_database}. Despite this difference, the testing split is kept consistent with the ICEBERG paper to maintain comparative evaluation; we also include an intermediate ablation (retrieval-free ICEBERG with collision energy) to fairly position the improvement of RAG. 
Since collision energies are found crucial to RAG and the NPLIB1 dataset used in \citet{goldman2024iceberg} does not have collision energy labels for most of the spectra, this dataset is not included for comparison.

We include all peer methods reported in the benchmark developed by \citet{goldman2024iceberg}.
Traditionally, MS/MS simulators have relied on combinatorial enumeration of bond breakages, which mimics the physical process, but they are computationally demanding and often inaccurate \citep{allen2015cfm-id, MAGMA}.
Deep learning-based simulators significantly reduce the inference time with competitive accuracy, while the physical constraint is either fully lifted~\citep{young2024massformer,wei2019neims} or relaxed to subformulae~\citep{murphy2023efficiently,goldman2023scarf}. ICEBERG combines both ideas by learning the fragmentation DAG. Besides them, we also consider a ``spectrum retrieval'' baseline by simply taking the retrieved spectrum as the prediction.

\begin{figure}[tb!]
\includegraphics[width=\linewidth]{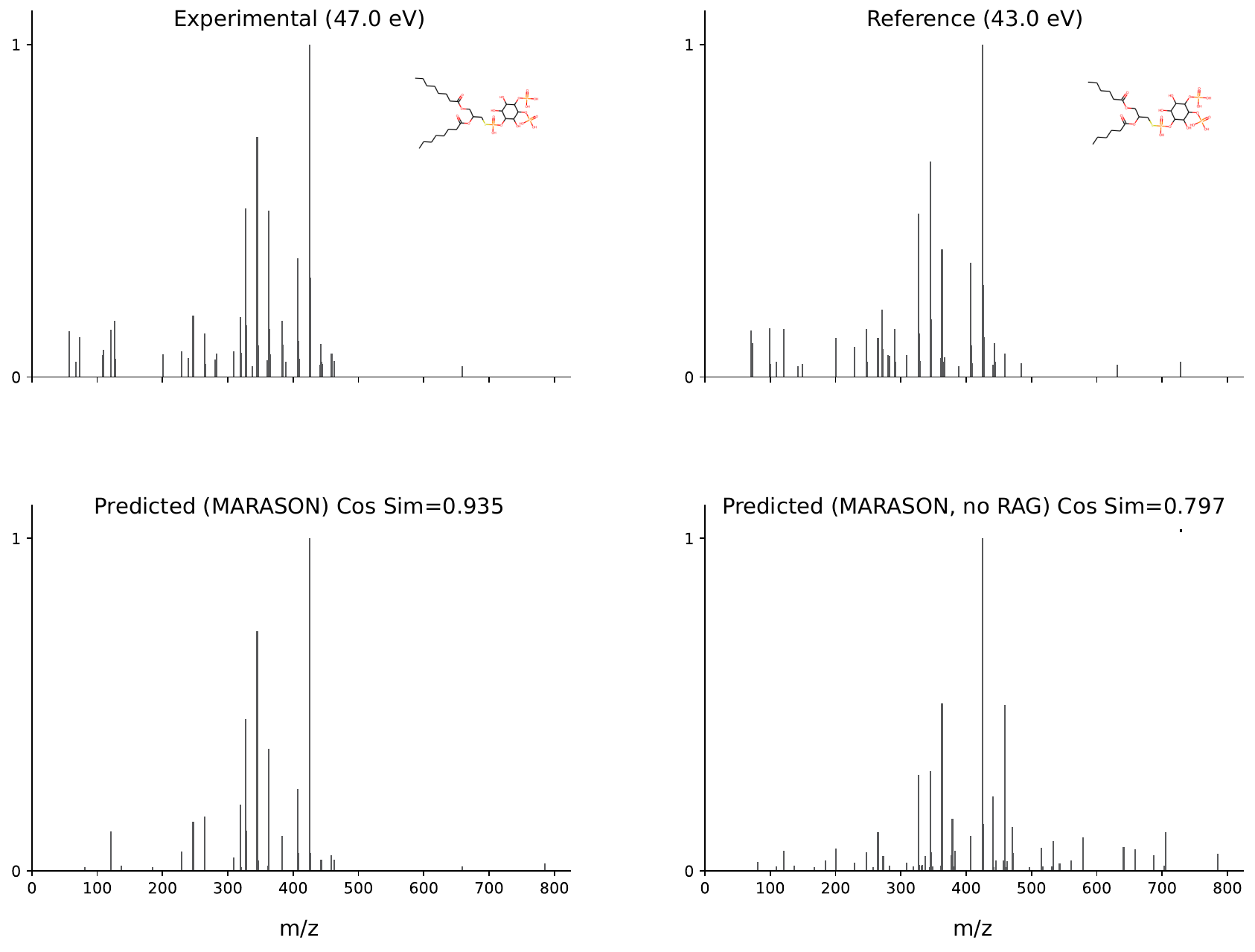}
    \vspace{-20pt}
    \caption{Qualitative comparison of experimental spectrum (i.e., ground-truth), reference spectrum, spectrum predicted by \ours, and spectrum predicted by \ours (no RAG). Cosine similarities between the experimental spectrum and predicted spectra are annotated. The retrieved reference structure, as shown in the plot, is structurally similar to the target structure, whereby the reference spectrum offers important intensity information. Our graph-matching-based RAG strategy associates reference peaks with the target, despite the peaks having different m/z values between the reference and the target.}    
    \vspace{-5pt}
    \label{fig:spec}
\end{figure}

\subsubsection{Quantitative Evaluation Metrics}
\textbf{Spectrum similarity}. We evaluate model performance via how similar the predicted mass spectra are to experimental spectra. Specifically, since we consider molecules under 1,500 Da, we create a 15,000-dimensional vector valued between 0 to 1 that encodes the spectrum, where each element means the peak intensity that falls within the mass bin of 0.1 Da, to balance mass resolution and the vector dimension sparsity. All method outputs are transformed into this 15,000-dim vector so that it accommodates both methods that predict binned spectra~\citep{wei2019neims,young2024massformer} and methods that predict intensities with known exact mass~\citep{murphy2023efficiently,goldman2023scarf,goldman2024iceberg}, including ours. In this paper, we use cosine similarity to compare spectra, which is the de facto standard metric used in real-world case studies~\citep{li2021entropy}.

\begin{table*}[!tb]
    \centering
    \caption{Retrieval accuracy (mean with
    99.9\% confidence intervals upon bootstrapping, 20,000 resamples) with known chemical formula on the MassSpecGym dataset~\citep{bushuiev2024massspecgym}. We consider MassSpecGym as a more challenging setting than \citet{nist_database} as it incorporates fewer annotated spectra and emphasizes generalization to different molecular scaffolds. \ours surpasses all baselines in terms of retrieval accuracy. Baseline performances are quoted from \citet{bushuiev2024massspecgym}.}
    \resizebox{0.8\linewidth}{!}
    {
    \begin{tabular}{l|ccc}
    \toprule
        Accuracy @ Top-$k$ & 1 & 5 & 20 \\ \midrule
        NEIMS (FFN)~\citep{wei2019neims} & 0.0762 (0.0677-0.0854) & 0.2270 (0.2132-0.2412) &  0.4412 (0.4251-0.4575) \\ 
        NEIMS (GNN)~\cite{zhu2020using} & 0.0363 (0.0305-0.0429) & 0.1355 (0.1246-0.1468) & 0.3377 (0.3226-0.3537) \\ 
        FraGNNet~\citep{young2024fragnnet} & 0.3193 (0.3040-0.3350) & 0.6320 (0.6164-0.6476) & 0.8270 (0.8145-0.8393) \\
        {\ours}~(ours) & \textbf{0.3403 (0.3286-0.3520)} & \textbf{0.6404 (0.6277-0.6519)} & \textbf{0.8539 (0.8448-0.8624)}\\
        \bottomrule
    \end{tabular}
    }
    \vspace{-5pt}
    \label{table:retrieval_msg}
\end{table*}

\textbf{Retrieval accuracy}. In structural elucidation campaigns, MS/MS simulators can predict pseudo-spectrum labels for candidate structures and compare them to the experimental spectrum. Candidate structures can then be ranked based on spectrum similarity to identify the most likely structure.
This retrieval setting evaluates how this model might be deployed in a real-world structural elucidation campaign. Following \citet{goldman2024iceberg}, for each molecule from the testing dataset, we obtain up to 49 isomeric decoys with the highest Tanimoto similarities to the true target structure, i.e., the decoys most likely to be mistaken under Tanimoto measurement. We run \ours and all comparative methods on 50 structures per test spectrum (49 decoys + 1 true structure) and rank all of them by cosine similarity. We evaluate the retrieval accuracy at top-$k$.

\begin{table}[!tb]
    \centering
    \caption{Ablation Study of MARASON design choices. RAG strategies include no RAG, concatenating one reference spectrum to the model input, traditional matching methods discussed in Sec.~\ref{sec:method-traditional-gm} including Hungarian~\citep{Kuhn1955Hungarian} and RRWM~\citep{ChoECCV10}, and neural graph matching (NGM) presented in Sec.~\ref{sec:method-ngm} where the matching layer could be either Sinkhorn~\citep{SinkhornAMS64} or Softmax. We compare cosine similarity on random split with seed $=1$.}
    \resizebox{\linewidth}{!}{
    \begin{tabular}{c|cc|c}
    \toprule
        Base model & RAG strategy & Match layer & Cosine sim. \\ \midrule
         & No RAG & - & 0.739~~~~~~~~~~~~~~ \\
         & Concat & - &  0.737 (-0.3\%) \\ \cmidrule{2-4}
         \ours & Hungarian & - & 0.746 (+0.9\%) \\ 
         (shared GNN) & RRWM & - & 0.742 (+0.4\%) \\ \cmidrule{2-4}
          & NGM & Sinkhorn & 0.749 (+1.4\%) \\ 
         & NGM & Softmax & 0.753 (+1.9\%) \\
        \midrule
        \ours & NGM & Sinkhorn & 0.753 (+1.9\%) \\ 
        (not shared GNN) & NGM & Softmax & \textbf{0.757 (+2.4\%)} \\ 
        \bottomrule
    \end{tabular}
    }
    \vspace{-5pt}
    \label{table:ablation}
\end{table}
\subsection{Results and Discussions}
\subsubsection{Spectral Similarity and Visualization}
We evaluate the prediction power of \ours by comparing the spectral cosine similarity of the predicted spectrum of different baseline models on the \citet{nist_database} dataset. The results are summarized in Fig.~\ref{fig:cosine}. \ours outperforms all baselines in both types of splits, and the success on scaffold split excludes the possibility that \ours needs highly similar reference structures for accurate predictions. It improves the cosine similarity of the ICEBERG baseline by a relative 5.2\% on random split and a relative 3.7\% on scaffold split. Since the original ICEBERG model does not consider collision energy, we also compare a collision energy-aware version of ICEBERG. This study demonstrates that neural graph matching-based RAG improves the performance of MS/MS simulation, a representative task in molecular machine learning. An example of predicted spectra of \ours is provided in Fig.~\ref{fig:spec}.

\subsubsection{Retrieval from PubChem Candidates}
The evaluation of the real-world applicability of MS/MS simulators is summarized in Table~\ref{table:retrieval}. In the retrieval benchmark, \ours improves the top-1 retrieval accuracy upon ICEBERG from 18.7\% to 27.8\%, a marked increase in the state-of-the-art performance for this task by a relative margin of 48\% over the next best method. 
As an ablation study, ICEBERG (with collision) has a top-1 accuracy of 20.2\%, indicating that the majority of improvement is directly attributable to our use of RAG. \ours consistently outperforms all baselines from top-1 to top-10 retrieval accuracies. 
Table~\ref{table:retrieval} only covers the random split; retrieval experiment results on scaffold split are in Table~\ref{tab:retrieval_scaffold} in the Appendix.

\subsubsection{Retrieval from the MassSpecGym dataset}
We further retrain \ours on the recently developed open-source dataset, MassSpecGym \cite{bushuiev2024massspecgym}, where we achieve state-of-the-art retrieval accuracy, as shown in Table~\ref{table:retrieval_msg}. Since not all spectra in MassSpecGym have collision energy label, but \ours requires labeled collision energies, we use a NIST-pretrained model to create pseudo labels for those unannotated spectra, i.e., transferring the collision energy knowledge from NIST to MassSpecGym. The model is retrained from scratch on MassSpecGym data to ensure no data leakage from NIST. \ours improves the top-1 retrieval accuracy upon the current state-of-the-art, FraGNNet, from 31.93\% to 34.03\%, a marked increase by a relative margin of 6\% over the next best method. Random seed is fixed as 1 for \ours.

\begin{figure*}[tb!]
\includegraphics[width=\linewidth]{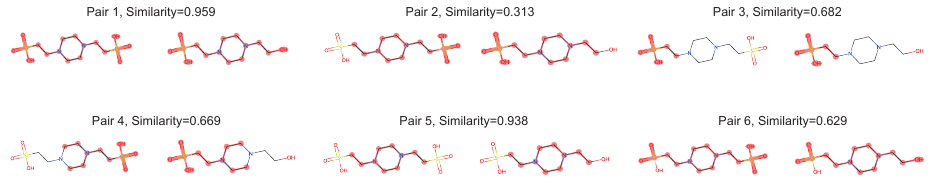}
    \vspace{-5pt}
    \caption{Visualization of matched fragment pairs and their Softmax-similarity scores discovered by the neural graph matching module in \ours. Structures on the left are from the target fragmentation DAG and structures on the right are from the reference. Fragments are highlighted from the original structure. Our matching module learns to match not only exact structures (pair 2, 3, 4, 5), but also correlated structures with slight modifications (pair 1, 6). Only 6 pairs from two fragmentation DAGs are shown due to space limits, the full list of matched pairs can be found in Fig.~\ref{fig:pairs} in the Appendix together with a visualization of traditional graph matching in Fig.~\ref{fig:pairs_rrwm}.}    
    \label{fig:pairs_partial}
\end{figure*}

\begin{table*}[tb!]
    \centering
    \caption{The performance of \ours with and without RAG on target and reference pairs grouped by Tanimoto similarity. We report cosine similarity on the random split with random seed = $1$. There is no testing sample with a Tanimoto similarity between 0 and 0.1.}
\resizebox{\linewidth}{!}{
    \begin{tabular}{l|cccccccccc}
    \toprule
        Tanimoto Similarity & (0, 0.1] & (0.1, 0.2] & (0.2, 0.3] & (0.3, 0.4] & (0.4, 0.5] & (0.5, 0.6] & (0.6, 0.7] & (0.7, 0.8] & (0.8, 0.9] & (0.9, 1.0] \\ \midrule
        \ours & N/A & 0.550 & 0.614 & 0.690 & 0.741 & 0.789 & 0.815 & 0.808 & 0.805 & 0.824 \\
        \ours (No RAG) & N/A & 0.566 & 0.611 & 0.682 & 0.727 & 0.768 & 0.791 & 0.780 & 0.759 & 0.784 \\
        \bottomrule
    \end{tabular}
    }
    \label{table:SimPerformance}
\end{table*}

\subsubsection{Ablation Study}
We conduct an ablation study to compare matching algorithms and GNN designs on the \citet{nist_database} dataset under a random split, as shown in Table~\ref{table:ablation}. \ours includes modest engineering changes to ICEBERG for more efficient training and inference on GPUs, which accounts for the slight improvement over ICEBERG (with collision). The naive RAG strategy that concatenates the reference spectrum as an extra 15,000-dimensional input leads to a \emph{negative} impact on the cosine similarity. Traditional graph matching methods discussed in Sec.~\ref{sec:method-traditional-gm}, by comparison, yield a minor benefit of fragment-level matching in RAG. With the comparatively more expressive and end-to-end learnable neural graph matching described in Sec.~\ref{sec:method-ngm}, the cosine similarity is further improved. 
A possible explanation for the superiority of Softmax over Sinkhorn is that Softmax is sufficient for the many-to-one aggregation path in Eq.~(\ref{eq:inten-set-trans-input}) and provides better gradients because it takes fewer iterations. It is also shown in \citet{sarlin2020superglue} that Softmax outperforms as the matching layer for larger-sized graphs. Finally, we discover that separating the GNN that learns embeddings for matching and the GNN that learns embeddings for intensity prediction results in a higher spectrum similarity, compared to using GNNs with shared weights for both purposes. The reason could be that \ours needs separate modules for learning intensity-related information (e.g., molecular fragment cross sections) and information about the fragmentation DAGs themselves.

\subsubsection{Qualitative Study of Matching Patterns}
We visualize matching pairs of fragments by assigning each target fragment to the reference fragment with the highest Softmax score.
An example of six matched fragment pairs is shown in Fig.~\ref{fig:pairs_partial}. 
Fragment pairs also show similar fragmentation patterns (missing the same functional group, C-N bond breakages, etc.). This example illustrates how \ours has learned to match fragments that are generated through similar mechanisms in the fragmentation process in order to model the relationship between each matched peak and fragment pair in the reference and target spectrum.
~\subsubsection{What if there is no reference structure?}
To understand the relationship between RAG's improvement and the availability of a similar-enough reference structure, for all testing data in \citet{nist_database}, we group structural pairs based on their Tanimoto similarities and compare performances of \ours with or without RAG (same model configurations as in Table~\ref{table:ablation}). As shown in Table~\ref{table:SimPerformance}, RAG starts to bring a significant performance gain when the Tanimoto similarity is larger than 0.3. When the best reference structure has a Tanimoto similarity between 0.1 and 0.3, RAG does not bring a significant improvement, while \ours still performs robustly with such irrelevant references. 
The slight performance drop between Tanimoto similarity (0.1, 0.2] also suggests a simple trick to further improve \ours's accuracy: use standard \ours when the retrieved structure has a Tanimoto similarity $> 0.2$ and use the non-RAG version otherwise.


\section{Conclusion}
This paper presents a retrieval-augmented generation framework for molecules by integrating neural graph matching into an existing end-to-end geometric deep learning framework. On the molecular machine learning task of mass spectrum simulation, we implement \ours to match the fragmentation DAGs of the target and reference structures, pair up peaks and fragments in the reference and target structures, and use that alignment information to generate improved spectrum predictions. \ours establishes new state-of-the-art performance in terms of the quality of simulated spectra and retrieval accuracy in downstream applications. Future improvements and adaptations could see this neural graph matching strategy applied to a broader range of retrieval-augmented structure-property prediction tasks across the field of molecular machine learning.

\section*{Acknowledgements}
This work was partly supported by DSO National Laboratories in Singapore and the MIT Undergraduate Research Opportunities Program (UROP).



\section*{Impact Statement}
This paper presents work that aims to advance both machine learning and mass spectroscopy. As a methodology-focused paper, this paper itself does not pose any significant societal impact that is required to be highlighted here. The authors are also aware that mass spectroscopy could be used in human-centric research campaigns such as metabolomics, and it is important to ensure that the reference database is not biased towards ethnicity, sex, and age, or at least being acknowledged about the potential bias. On the other hand, efforts to develop better mass spectroscopy models, including this paper, will potentially lower the cost and expedite those discovery campaigns. If such models are used properly for the underrepresented cohort, it will finally mitigate the bias and unfairness in human-centric studies.





\bibliography{example_paper}
\bibliographystyle{icml2025}


\newpage
\appendix
\onecolumn

\section{Retrieval Accuracy on Scaffold Split}

Following most peer methods, the retrieval accuracy evaluation in the main paper (Table~\ref{table:retrieval}) mainly focuses on random split and all positive adduct types. We also include the retrieval accuracy on scaffold split as follows. Scaffold split is considered more challenging than random split as it separates structures with different molecular scaffolds into different training or testing sets, requiring the model with more generalization ability for out-of-distribution structures. \ours maintains its performance superiority over the baselines on scaffold split. Since scaffold split is less studied in peer methods there are fewer baselines in Table~\ref{table:retrieval}. Multiple random restarts are not considered here because the standard deviation in random split is relatively small, indicating that most peer methods are somewhat stable against random seeds. 

\begin{table*}[h]
    \centering
    \caption{Retrieval Accuracy upon NIST20 Dataset with scaffold split on positive adduct types.}
    \begin{tabular}{l|ccccccc}
    \toprule
        Accuracy @ Top-$k$ & 1 & 2 & 3 & 4 & 5 & 8 & 10 \\ \midrule
        Graff-MS~\citep{murphy2023efficiently} & 0.142 & 0.265 & 0.36 & 0.446 & 0.508 & 0.636 & 0.703 \\   MassFormer~\citep{young2024massformer} & 0.178 & 0.318 & 0.422 & 0.506 & 0.568 & 0.706 & 0.768 \\ \midrule   ICEBERG~\citep{goldman2024iceberg} & 0.206 & 0.396 & 0.519 & 0.604 & 0.658 & 0.769 & 0.815 \\ 
        MARASON (ours) & 0.283 & 0.464 & 0.567 & 0.645 & 0.700 & 0.814 & 0.853 \\
        \bottomrule
    \end{tabular}
    \label{tab:retrieval_scaffold}
\end{table*}

\section{Retrieval Accuracy with [M+H]+ Only}

One design choice in MS/MS simulator development is how many adduct types are supported by the model. Our \ours aims to cover most adduct types in the \citet{nist_database} database, while some methods e.g.\ FraGNNet~\citep{young2024fragnnet} only support the most common adduct type---[M+H]+. It is still feasible to compare these methods by restricting testing adduct type as [M+H]+, where the retrieval accuracy is shown as follows in Table~\ref{table:retrieval_H}. The performance of FraGNNet is quoted from their paper (therefore missing accuracies at $k=2,4,8$, also missing error bars), and other baselines are implemented with our benchmark. It is worth noting that although \citet{young2024fragnnet} report better retrieval accuracy of FraGNNet than all peer methods, our reevaluation of all methods on the [M+H]+-only subset shows that ICEBERG~\citep{goldman2024iceberg} still outperforms, suggesting that single-bond breaking in MS/MS simulator design might not be superior to multiple-bond breaking. Incorporating collision energy for ICEBERG also improves the retrieval accuracy on [M+H]+, and our \ours reaches state-of-the-art retrieval accuracy.

\begin{table*}[h]
    \centering
    \caption{Retrieval Accuracy (mean $\pm1.96$ standard deviation of three random seeds) upon \citet{nist_database} dataset with random (InChI key) split on the [M+H]+ adduct type. }
    \resizebox{\linewidth}{!}{
    \begin{tabular}{l|ccccccc}
    \toprule
        Accuracy @ Top-$k$ & 1 & 2 & 3 & 4 & 5 & 8 & 10 \\ \midrule
        Graff-MS~\citep{murphy2023efficiently} & 0.211$\pm$0.004 & 0.365$\pm$0.009 & 0.472$\pm$0.015 & 0.551$\pm$0.013 & 0.608$\pm$0.005 & 0.723$\pm$0.008 & 0.775$\pm$0.009 \\  MassFormer~\citep{young2024massformer} & 0.252$\pm$0.001 & 0.422$\pm$0.002 & 0.539$\pm$0.005 & 0.617$\pm$0.007 & 0.675$\pm$0.004 & 0.794$\pm$0.010 & 0.843$\pm$0.006 \\   FraGNNet~\citep{young2024fragnnet} & 0.238 & - & 0.504 & - & 0.652 & - & 0.831\\ 
        \midrule
        ICEBERG~\citep{goldman2024iceberg} & 0.251$\pm$0.016 & 0.454$\pm$0.004 & 0.576$\pm$0.0060.718 & 0.654$\pm$0.004 & 0.711$\pm$0.007 & 0.810$\pm$0.001 & 0.850$\pm$0.009\\ 
        ICEBERG (w/ collision energy) & 0.270$\pm$0.016 & 0.487$\pm$0.009 & 0.611$\pm$0.011 & 0.679$\pm$0.013 & 0.735$\pm$0.013 & 0.840$\pm$0.009 & 0.877$\pm$0.008\\ 
        \textbf{{MARASON} (ours)} & \textbf{0.331$\pm$0.002}& \textbf{0.520$\pm$0.005} & \textbf{0.633$\pm$0.010} & \textbf{0.706$\pm$0.005} & \textbf{0.754$\pm$0.002} & \textbf{0.849$\pm$0.003} & \textbf{0.885$\pm$0.003}  \\ \hline
    \bottomrule
    \end{tabular}
    }
    \label{table:retrieval_H}
\end{table*}
\newpage
\section{Cosine Similarity}
We include the detailed cosine similarity numbers from Fig.~\ref{fig:cosine} for better reproducibility in future works. 
\begin{table}[h]
    \caption{Experimental results in line with Fig.~\ref{fig:cosine} of cosine similarity. Random split has mean $\pm1.96$ standard deviation across 3 random seeds.}
    \centering
    \begin{tabular}{l|cc}
    \toprule
        Models & Random & Scaffold \\
        \midrule
        Retrieved Spec & 0.359 & 0.334 \\ 
        \midrule
        3DMolMS~\citep{hong20233dmolms} & 0.51$\pm$0.001 & 0.466 \\ 
        NEIMS (FFN)~\citep{wei2019neims} & 0.617$\pm$0.001 & 0.546 \\ 
        NEIMS (GNN)~\citep{zhu2020using} & 0.694$\pm$0.001 & 0.643 \\ 
        FixedVocab~\citep{murphy2023efficiently} & 0.704$\pm$0.001 & 0.658 \\ 
        MassFormer~\citep{young2024massformer} & 0.721$\pm$0.004 & 0.682 \\ 
        SCARF~\citep{goldman2023scarf} & 0.726$\pm$0.002 & 0.669 \\
        \midrule
        ICEBERG~\citep{goldman2024iceberg} & 0.727$\pm$0.002 & 0.699 \\ 
        ICEBERG (w/ colli.) & 0.735$\pm$0.005 & 0.711 \\ \midrule
        \textbf{\ours (ours)} & \textbf{0.765$\pm$0.002} & \textbf{0.725} \\ \hline
        \bottomrule
    \end{tabular}
    \label{table:cosine}
\end{table}

\section{Evaluation of Matching Patterns: Visualizing Failure Modes of Traditional Graph Matching}

\begin{figure*}[tb!]
    \centering
\includegraphics[width=\linewidth]{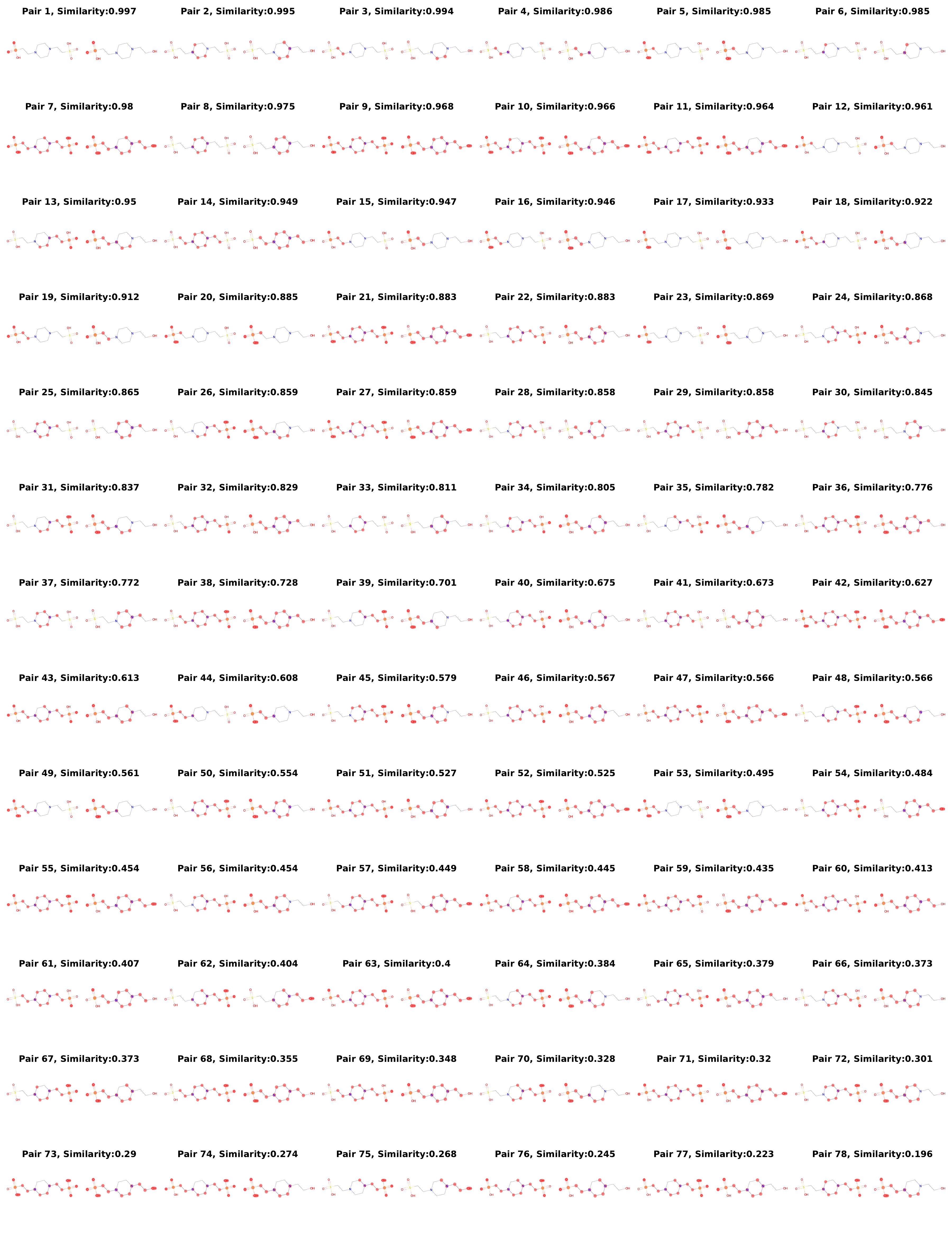}
    \vspace{-30pt}
    \caption{A full list of fragment pairs generated by \ours, sorted by the learned Softmax scores. Structures on the left are from the target fragmentation DAG and structures on the right are from the reference.}\label{fig:pairs}
\end{figure*}

For the qualitative analysis of matching pairs identified by \ours, Fig.~\ref{fig:pairs_partial} in the main text visualizes only six fragment pairs due to space constraints. Here, ICEBERG-Generate predicts 78 fragments for the target molecule, and we provide a complete list of all 78 matched fragment pairs. As shown in Fig.~\ref{fig:pairs}, the original target structure (left) and the original reference structure (right) are included, with fragments highlighted by their respective atoms and bonds.  

In Fig.~\ref{fig:pairs}, the top 50 matched pairs, which exhibit higher similarity scores, predominantly correspond to exact structural matches. In contrast, the lower-scored pairs include more fragments that, while structurally distinct, follow similar fragmentation pathways. Notably, the learned matching function in \ours effectively identifies an informative and chemically interpretable fragment-matching strategy. This ensures that the retrieved intensity information is accurately assigned to the correct target fragments, thereby enhancing MS/MS simulation accuracy.  

\begin{figure*}[tb!]
    \centering
\includegraphics[width=\linewidth]{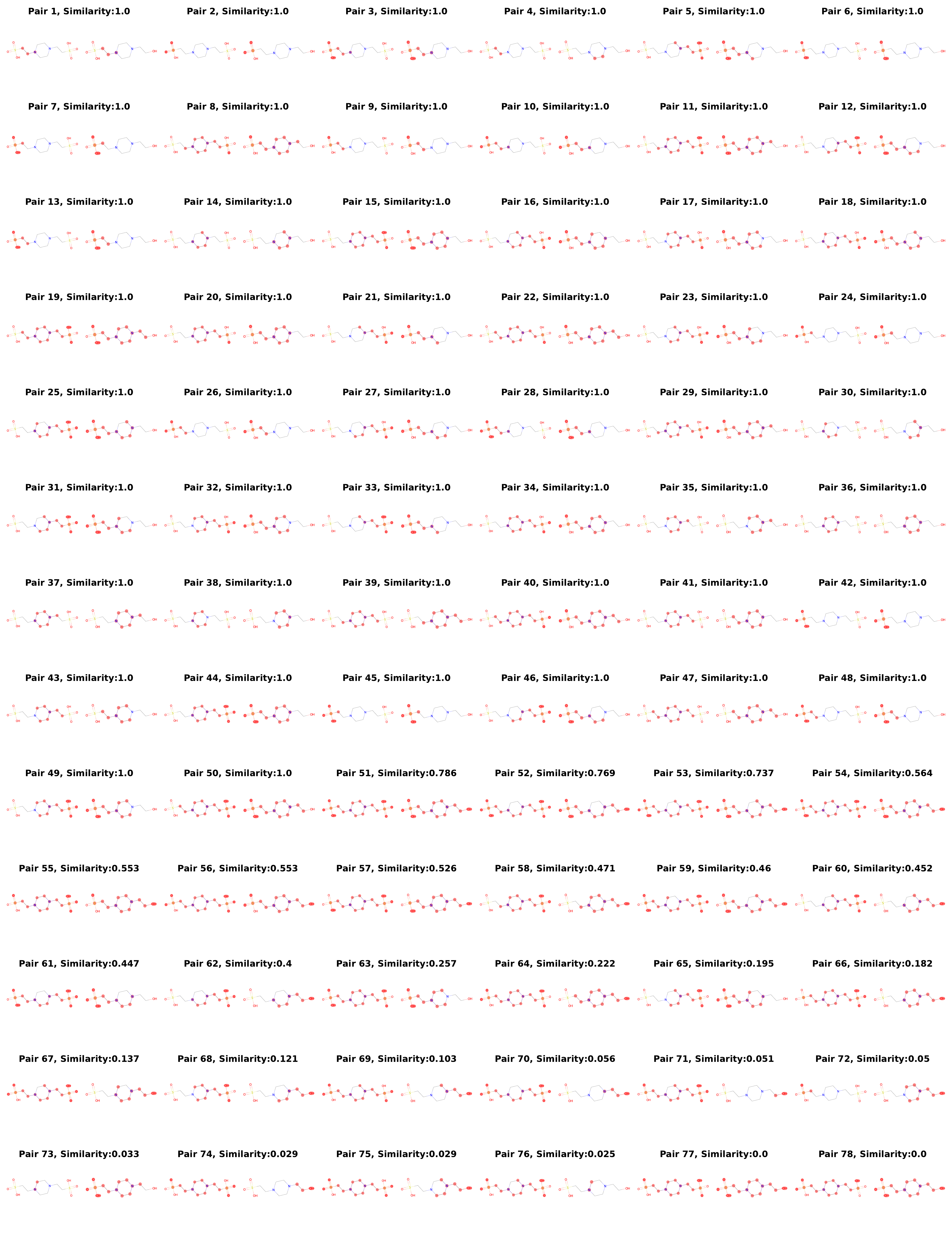}
    \vspace{-30pt}
    \caption{A full list of fragment pairs generated by RRWM graph matching~\citep{ChoECCV10}, sorted by fixed Tanimoto similarity scores. Structures on the left are from the target fragmentation DAG and structures on the right are from the reference.}\label{fig:pairs_rrwm}
\end{figure*}

For matching pairs identified by RRWM (Reweighted Random Walk Matching)~\citep{ChoECCV10}, which relies on fixed graph-matching affinity metrics, a greater number of unreasonable matchings occur, particularly when the fragment structures are not identical (see Fig.~\ref{fig:pairs_rrwm}). This issue is especially pronounced for pairs 65–78, where the similarity score falls below 0.2. The limitation of traditional graph matching arises from its fixed affinity metric: for identical fragments, it functions effectively, producing accurate matches; however, for non-identical fragments, it assigns noisy similarity scores, often categorizing them as matching ``outliers'' (fragments with low affinity scores to all others). As a result, these fragments are incorrectly matched with unrelated structures.  

Additionally, within the tail distribution of challenging fragment pairs, the learned neural graph-matching method yields higher similarity scores, demonstrating its superior ability to capture meaningful fragment correspondences.


\end{document}